\documentclass[conference]{IEEEtran}
\IEEEoverridecommandlockouts
\usepackage{cite}
\usepackage{amsmath,amssymb,amsfonts}
\usepackage{graphicx}
\usepackage{textcomp}
\usepackage{xcolor}
\usepackage{url}
\usepackage{algorithm}
\usepackage{algpseudocode}
\usepackage{float}

\usepackage{xcolor}
\usepackage{listings}
\lstdefinestyle{py}{
  language=Python,
  basicstyle=\ttfamily\small,
  keywordstyle=\color{blue}\bfseries,
  commentstyle=\color{gray}\itshape,
  stringstyle=\color{green!50!black},
  frame=single,
  breaklines=true,
  tabsize=2,
  showstringspaces=false,
  captionpos=b
}
\def\BibTeX{{\rm B\kern-.05em{\sc i\kern-.025em b}\kern-.08em
    T\kern-.1667em\lower.7ex\hbox{E}\kern-.125emX}}
\begin{document}

\title{
    Adaptive Policy Synchronization for Scalable Reinforcement Learning
    
    \thanks{Research was sponsored by the Department of the Air Force Artificial Intelligence Accelerator and was accomplished under Cooperative Agreement Number FA8750-19-2-1000. The views and conclusions contained in this document are those of the authors and should not be interpreted as representing the official policies, either expressed or implied, of the Department of the Air Force or the U.S. Government. The U.S. Government is authorized to reproduce and distribute reprints for Government purposes notwithstanding any copyright notation herein.}
}

\author{\IEEEauthorblockN{Rodney Lafuente-Mercado}
\IEEEauthorblockA{\textit{MIT Lincoln Laboratory} \\
Lexington, MA \\
Rodney.LafuenteMercado@ll.mit.edu}
}

\maketitle

\begin{abstract}
Scaling reinforcement learning (RL) often requires running environments across many machines, but most frameworks tie simulation, training, and infrastructure into rigid systems. We introduce \textbf{ClusterEnv}, a lightweight interface for distributed environment execution that preserves the familiar Gymnasium API. ClusterEnv uses the \textbf{DETACH} pattern, which moves environment \texttt{reset()} and \texttt{step()} operations to remote workers while keeping learning centralized. To reduce policy staleness without heavy communication, we propose \textbf{Adaptive Policy Synchronization} (APS), where workers request updates only when divergence from the central learner grows too large. ClusterEnv supports both on- and off-policy methods, integrates into existing training code with minimal changes, and runs efficiently on clusters. Experiments on discrete control tasks show that APS maintains performance while cutting synchronization overhead. Source code is available at \url{https://github.com/rodlaf/ClusterEnv}
.
\end{abstract}


\section{Introduction}

Distributed reinforcement learning (DRL) has become a central paradigm for scaling training across large compute clusters \cite{arulkumaran_deep_2017}\cite{liu_acceleration_2024}. However, most existing DRL frameworks tightly couple environment execution with learning logic, replay management, and orchestration infrastructure. This coupling introduces rigidity: users must adopt a complete framework stack just to distribute simulation, even when they wish to retain their own training pipelines. This friction is particularly pronounced in research and production settings where modularity, experimentation, and system-level customization are critical.

We introduce ClusterEnv, a lightweight interface for distributed environment execution that isolates only the simulation layer. Inspired by the design of OpenAI Gym and its successor Gymnasium, ClusterEnv provides a drop-in environment replacement that supports remote execution of \texttt{reset()} and \texttt{step()} operations across a compute cluster. Training code, model definitions, and optimization logic remain entirely centralized and under user control. This makes ClusterEnv compatible with arbitrary learning algorithms, including on-policy and off-policy methods, and easily integrable with lightweight libraries such as CleanRL.

ClusterEnv follows the DETACH architecture: \textit{Distributed Environment execution with Training Abstraction and Centralized Head}. Workers handle only environment interaction and local inference, while the head node maintains the learner state and training loop. This two-tiered structure avoids complex synchronization logic, parameter servers, or entangled dataflows, enabling simpler rollout collection at scale.

A central challenge in distributed systems is \textit{policy staleness}—the drift that arises when remote actors collect data using outdated versions of the policy. Traditional systems often rely on fixed interval broadcasting or post-hoc corrections such as importance sampling to address this issue. Instead, we propose Adaptive Policy Synchronization (APS), a mechanism in which each actor monitors its own policy divergence from the central learner and requests updates only when this divergence exceeds a configurable threshold. This strategy ensures bandwidth-efficient synchronization without compromising sample utility or requiring algorithm-specific corrections.

Together, ClusterEnv, DETACH, and APS provide a unified system-level contribution: a learner-agnostic, high-throughput, and modular approach to distributed DRL. We validate our design on classic discrete control tasks using Proximal Policy Optimization (PPO), a widely-used on-policy method. With this experiment we demonstrate that APS achieves strong performance while substantially reducing synchronization overhead. The code is publicly available at \url{https://github.com/rodlaf/ClusterEnv}.

\noindent\textbf{Contributions.} (1) We formalize \emph{DETACH}, a two-tier design that isolates distributed simulation from learning and orchestration, enabling drop-in use with existing training loops. (2) We propose \emph{Adaptive Policy Synchronization} (APS), a divergence-triggered policy refresh that reduces bandwidth without algorithm-specific corrections. (3) We provide an open-source implementation with SLURM integration and a Gymnasium-compatible API.

\section{Related Work}

The use of distributed computation in deep reinforcement learning (DRL) traces back to early experiments with Deep Q-Networks (DQN) on Atari games \cite{mnih_playing_2013}, where training speed and sample throughput were primary bottlenecks. The first formal distributed DRL architecture, General Reinforcement Learning Architecture (GORILA)\cite{nair_massively_2015}, introduced a centralized design featuring four components: actors for data collection, a parameter server for model updates, learners for gradient computation, and a shared replay memory. This design enabled DQN to scale across machines by coordinating asynchronous updates from actors and learners.

APE-X \cite{horgan_distributed_2018} built upon GORILA by simplifying the architecture and improving sample efficiency. It retained the centralized design but consolidated the parameter server and learner into a single GPU node, allowing for more efficient updates. Actors performed local inference with stale parameters and sent prioritized experiences to a central buffer. The introduction of a prioritized replay buffer, which stores past experiences for off-policy learning and ranks them by temporal-difference (TD) error, improved sample quality and accelerated convergence. APE-X was also extended to support other off-policy algorithms such as Deep Deterministic Policy Gradient (DDPG) \cite{lillicrap_continuous_2016}.

 Asynchronous Advantage Actor Critic (A3C) \cite{mnih_asynchronous_2016} took a different approach by decentralizing both data collection and learning. Multiple actor-learners asynchronously interacted with environments and updated shared model parameters using their own computed gradients. This eliminated the need for experience replay and enabled on-policy updates, but also introduced instability due to asynchronous gradient application and limited scalability.

The Importance Weighted Actor-Learner
Architecture (IMPALA) \cite{espeholt_impala_2018} reintroduced a centralized learner architecture while scaling simulation throughput via CPU-based actors. Actors collected rollouts using a stale snapshot of the policy and sent trajectories to a centralized GPU learner. The key innovation of IMPALA was the V-trace off-policy correction algorithm, which accounted for policy lag between actor and learner. V-trace is an off-policy correction algorithm that adjusts value targets using truncated importance, a statistical technique to correct for distribution mismatch, ensuring stable learning when actors collect data with stale policies.

\subsection{IMPALA and the Role of V-trace}

IMPALA decouples environment simulation from learning. Actors generate trajectories using stale policies and asynchronously send them to the central learner. To mitigate the instability introduced by policy lag, V-trace applies truncated importance sampling corrections.

Formally, let $\mu$ denote the behavior policy used by an actor and $\pi$ the target (learner) policy. At time step $t$, the actor observes a state $s_t$ and selects an action $a_t$. The V-trace correction weights are defined as

\[
\rho_t = \frac{\pi(a_t \mid s_t)}{\mu(a_t \mid s_t)}, \quad c_t = \min(\bar{c}, \rho_t),
\]

where $\rho_t$ is the importance weight and $c_t$ is a clipped version of that weight with truncation level $\bar{c}$. These weights are then used to compute corrected value targets, ensuring stable convergence despite the asynchronous architecture. This enables IMPALA to scale to large clusters without compromising learning stability.

\subsection{Environment Abstractions and Software Systems}

Several open-source systems provide abstractions for distributed reinforcement learning. RLlib \cite{liang_rllib_2018}, built on Ray \cite{moritz_ray_2018}, offers a high-level API supporting a wide range of RL algorithms. It abstracts both vectorized environments and distributed training but introduces significant framework overhead and tight integration across components. Similarly, Tianshou \cite{weng_tianshou_2022} is a modular framework with built-in distributed support, emphasizing reproducibility and algorithmic transparency.

EnvPool \cite{weng_envpool_2022}, in contrast, focuses on single-machine parallelism. Implemented in C++ with Python bindings, it achieves extremely high throughput for vectorized environments but is not designed for multi-node distributed execution.

Despite their flexibility, these libraries often enforce tightly coupled design patterns where environment execution, sampling, and learning are co-dependent. This creates friction for users looking to substitute only parts of the stack—for instance, replacing only the environment layer while retaining custom learning pipelines.

\subsection{DETACH and Adaptive Policy Synchronization}

While existing distributed reinforcement learning frameworks like IMPALA and APE-X offer scalable architectures, they tend to impose rigid roles on components—such as centralized learners, replay buffers, or scheduled weight broadcasting—thus limiting interoperability with custom training pipelines. Moreover, policy synchronization mechanisms in these systems are often fixed or entangled with the learning algorithm, making them hard to replace or reconfigure without modifying core internals.

We address these limitations by introducing the DETACH architecture. DETACH decouples the simulation layer from the learning process, allowing remote workers to run environments independently while all learning logic remains centralized and untouched. Unlike GORILA or IMPALA, which define distinct roles for actors, learners, and replay buffers, DETACH collapses coordination into a simpler two-part structure: centralized head and distributed workers. This makes it easy to swap in arbitrary training code—including model-free or model-based methods—without being forced into a prescriptive pipeline.

One key challenge in centralized architectures is managing the staleness of policies used by remote workers. Prior systems often address this with periodic weight broadcasting or post-hoc corrections (e.g., importance sampling in V-trace). In contrast, we propose an adaptive synchronization strategy that dynamically regulates when workers update their local policies based on observed divergence from the central learner. This method maintains the benefits of offloaded simulation while avoiding the high cost or instability that can arise from fixed update schedules or stale trajectories.

The result is a highly modular, bandwidth-efficient pattern that preserves the generality of existing RL implementations while enabling seamless scale-out. A detailed description of the DETACH architecture and its adaptive synchronization mechanism is provided in the next section.

\section{Methods}

In this section, we detail the system-level design and implementation of ClusterEnv, a lightweight, learner-agnostic interface for distributed reinforcement learning (RL) simulation. ClusterEnv is architected around a minimal but powerful abstraction that separates environment simulation from learning, while preserving compatibility with the Gymnasium API \cite{towers_gymnasium_2024}. At the heart of our design is the DETACH pattern (Distributed Environment execution with Training Abstraction and Centralized Head), which decentralizes rollout collection while centralizing all learning. We also introduce \textbf{Adaptive Policy Synchronization (APS)}, a principled mechanism for minimizing policy staleness without fixed synchronization intervals.

\subsection{The DETACH Architecture}

\begin{figure}[H]
    \centering
    \includegraphics[width=0.95\linewidth]{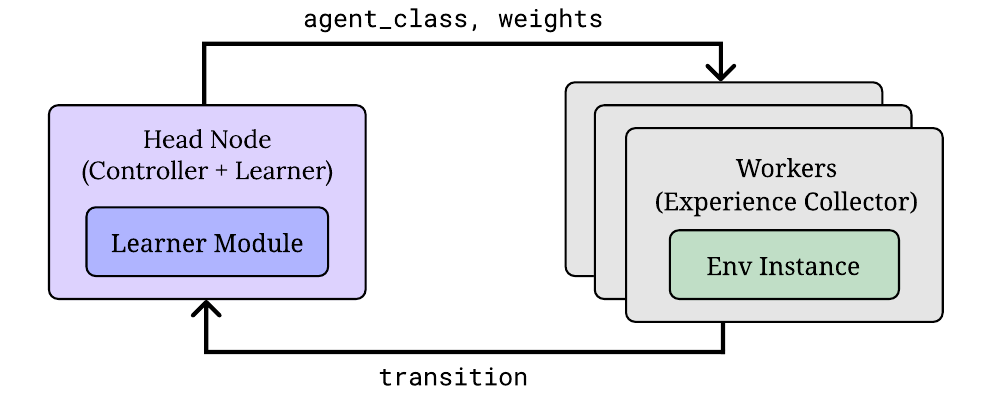}
    \caption{\textbf{The DETACH architecture.} Environment simulation is offloaded to distributed workers, while learning remains centralized at the head node. This separation ensures modularity and scalability without imposing rigid synchronization protocols.}
    \label{fig:detach-architecture}
\end{figure}

ClusterEnv implements a centralized distributed architecture that mirrors but simplifies systems such as APE-X and IMPALA. In this paradigm, all training logic—including model updates, gradient descent, and policy storage—resides in a centralized head node. Meanwhile, environment simulation is offloaded to remote worker nodes that execute only the \texttt{reset()} and \texttt{step()} methods.

This design pattern, which we refer to as DETACH, prioritizes system modularity and conceptual clarity. The environment workers perform local policy inference and communicate only observations, actions, and rewards with the learner. This eliminates the need for parameter servers or replay buffers tied into the environment, thereby enabling compatibility with arbitrary learners, optimizers, or loss functions.

To mitigate the issue of policy drift—where remote actors operate on stale policy parameters—we integrate APS directly into the ClusterEnv runtime. APS allows each worker to monitor the divergence between its local policy and the learner’s most recent policy and request weight updates only when necessary. This yields high communication efficiency while keeping trajectories sufficiently on-policy.

\subsection{Gymnasium-Compatible API}

ClusterEnv conforms closely to the standard Gymnasium environment interface, extending only minimally to support distributed execution. Specifically, the \texttt{ClusterEnv} class supports the following methods:

\begin{itemize}
    \item \texttt{reset()}: initializes environments across remote workers.
    \item \texttt{step(agent)}: performs a single environment step on all workers using the agent's policy.
\end{itemize}

To enable local inference on remote workers, we introduce a minimal extension to the \texttt{step()} interface: the agent is passed explicitly into the function. This agent must expose two methods:

\begin{itemize}
    \item \texttt{act(obs)}: returns an action given an observation.
    \item \texttt{get\_parameters()}: returns the agent’s current parameters for synchronization.
\end{itemize}

This abstraction preserves compatibility with Gym-style training loops, requiring only minimal modification for distributed execution.

\subsection{Adaptive Policy Synchronization (APS)}

One challenge in distributed RL systems is balancing communication efficiency with the fidelity of policy execution. Traditional architectures like APE-X or IMPALA either broadcast updated weights at fixed intervals or apply post-hoc corrections (e.g., V-trace) to compensate for stale data. Instead, we introduce APS, which adopts a proactive, divergence-aware strategy.

Each actor maintains a local policy $\pi_{\theta'}$ and periodically compares it to the latest central learner policy $\pi_\theta$. The KL divergence between the two policies at timestep $t$ for state $s_t$ is given by:

\[
D_{\text{KL}}(\pi_{\theta'}(\cdot \mid s_t) \,\|\, \pi_\theta(\cdot \mid s_t)) = \sum_{a \in \mathcal{A}} \pi_{\theta'}(a \mid s_t) \log \frac{\pi_{\theta'}(a \mid s_t)}{\pi_\theta(a \mid s_t)}.
\]

If the running average of this divergence, estimated over recent observations, exceeds a user-defined threshold $\delta > 0$, the actor requests updated parameters from the head node:

\[
\text{If} \quad \mathbb{E}_{s \sim \mathcal{D}}[D_{\text{KL}}(\pi_{\theta'}(\cdot \mid s) \,\|\, \pi_\theta(\cdot \mid s))] > \delta \quad \text{then} \quad \theta' \leftarrow \theta.
\]

This strategy ensures that workers remain within a bounded divergence from the learner’s policy, maintaining sample utility without incurring unnecessary synchronization costs. Notably, APS does not require changes to the training algorithm or policy loss and is compatible with both on-policy and off-policy methods.

\subsection{Pseudocode for the DETACH Pattern}

We now present pseudocode that formalizes the DETACH pattern, consisting of a centralized controller (head node) and distributed experience collectors (worker nodes). Synchronization is initiated dynamically based on APS divergence estimates.

\begin{algorithm}[H]
\caption{Head Node control loop (DETACH)}
\label{alg:detach-head}
\begin{algorithmic}
\State \textbf{Head Node (Controller)}
\State Initialize environment interface $\mathcal{E}$, learner policy $\pi$
\While{training not converged}
    \ForAll{workers $\text{worker}_i$}
        \State Send \texttt{Reset()} to $\text{worker}_i$
    \EndFor
    \While{episode not terminated}
        \ForAll{messages from workers}
            \If{message = \texttt{WeightRequest}}
                \State Send $\pi.\texttt{get\_parameters()}$ to worker
            \ElsIf{message = transition $(\mathbf{o}, a, r, \delta)$}
                \State Send transition to learner
            \EndIf
        \EndFor
    \EndWhile
\EndWhile
\end{algorithmic}
\end{algorithm}

\begin{algorithm}[H]
\caption{Worker Node execution loop (DETACH)}
\label{alg:detach-worker}
\begin{algorithmic}
\State \textbf{Worker Node (Experience Collector)}
\State Instantiate env $\mathcal{M}$, local policy $\pi_{\text{actor}}$
\While{active}
    \If{$\text{KL}(\pi_{\text{actor}}, \pi_{\text{cached}}) > \delta$}
        \State Send \texttt{WeightRequest} to head
        \State $\pi_{\text{actor}} \gets$ received weights
        \State $\pi_{\text{cached}} \gets \pi_{\text{actor}}$
    \EndIf
    \If{received \texttt{Reset()}}
        \State $\mathbf{o} \gets \mathcal{M}.\texttt{reset()}$
    \Else
        \State $a \gets \pi_{\text{actor}}(\mathbf{o})$
        \State $(\mathbf{o}', r, \delta, \xi) \gets \mathcal{M}.\texttt{step}(a)$
        \State Send $(\mathbf{o}, a, r, \delta)$ to head
        \State $\mathbf{o} \gets \mathbf{o}'$
    \EndIf
\EndWhile
\end{algorithmic}
\end{algorithm}

\subsection{Programmer-Centric Interface and SLURM Integration}

ClusterEnv is designed for conceptual economy: users write conventional single-node RL code and seamlessly gain distributed simulation. The following example shows how a programmer can run distributed rollouts using SLURM, a widely used HPC workload manager, while keeping the training loop simple and portable:

\begin{figure}[H]
\centering
\begin{lstlisting}[language=Python, caption={Distributed rollout collection using \texttt{ClusterEnv}.}, label={lst:usage}]
from clusterenv import ClusterEnv, SlurmConfig

env = ClusterEnv(
    env_config={
        "type": "gymnasium", 
        "kl_threshold": args.kl_threshold,
        "env_name": args.env_id, 
        "envs_per_node": args.envs_per_node,
    },
    config=SlurmConfig(
        job_name="lunar_lander",
        nodes=args.num_nodes,
        gpus_per_node=2,
        partition="gpu"
    )
)
env.launch()

obs = env.reset()
for _ in range(1000):
    obs, reward, done, info = env.step(agent)
\end{lstlisting}
\end{figure}

All orchestration, communication, and divergence tracking are handled internally. The user maintains control over the learning loop and is free to use any reinforcement learning library or model implementation.

\subsection{Summary}

The methods introduced here combine architectural simplicity with strong modularity. By decoupling rollout collection from learning, supporting Gym-style APIs, and introducing a lightweight divergence-aware synchronization mechanism (APS), ClusterEnv enables scalable distributed RL without entangling users in heavy framework dependencies. The DETACH pattern ensures clean separation of concerns, while APS provides dynamic, bandwidth-efficient synchronization that generalizes across algorithms and training setups.

\section{Experiments}

We use CleanRL~\cite{huang_cleanrl_2022}, a minimalist and reproducible RL library, as our training framework and pair it with ClusterEnv for distributed environment execution. All experiments use Proximal Policy Optimization (PPO) \cite{schulman_proximal_2017}, a standard on-policy training method, on the classic discrete-control benchmark \texttt{LunarLander-v2}.

\subsection{Setup}

ClusterEnv is configured with 4 nodes, each running 64 vectorized environments, for a total of 256 parallel environments. The PPO agent is trained for 5 million timesteps using the following hyperparameters:

\begin{itemize}
    \item \textbf{Environment:} \texttt{LunarLander-v2}
    \item \textbf{Timesteps:} 5 million
    \item \textbf{Learning rate:} $5 \times 10^{-4}$
    \item \textbf{Steps per rollout:} 1024
    \item \textbf{Mini-batches:} 8
    \item \textbf{Update epochs:} 30
    \item \textbf{Discount factor:} $\gamma = 0.99$
    \item \textbf{GAE lambda:} $\lambda = 0.95$
\end{itemize}

We evaluate the performance of our proposed \textbf{Adaptive Policy Synchronization} (APS) method by varying the KL divergence threshold $\delta$, which controls how tolerant actors are to policy drift before synchronizing with the learner.

\subsection{Ablation Over KL Thresholds}

We sweep over eight values of the KL divergence threshold:

\[
\delta \in \{ 0.001, 0.01, 0.05, 0.1, 0.2, 0.5, 0.8, 1.0 \}
\]

A lower threshold leads to more frequent policy synchronizations (more on-policy behavior), while higher thresholds allow more policy drift and reduce communication overhead.

\subsection{Results}

We report episode return averaged over evaluation rollouts as training progresses. Additionally, we track the number of policy synchronization events per worker, logged via TensorBoard, to measure the efficiency of APS.

\begin{figure}[ht]
    \centering
    \includegraphics[width=0.95\linewidth]{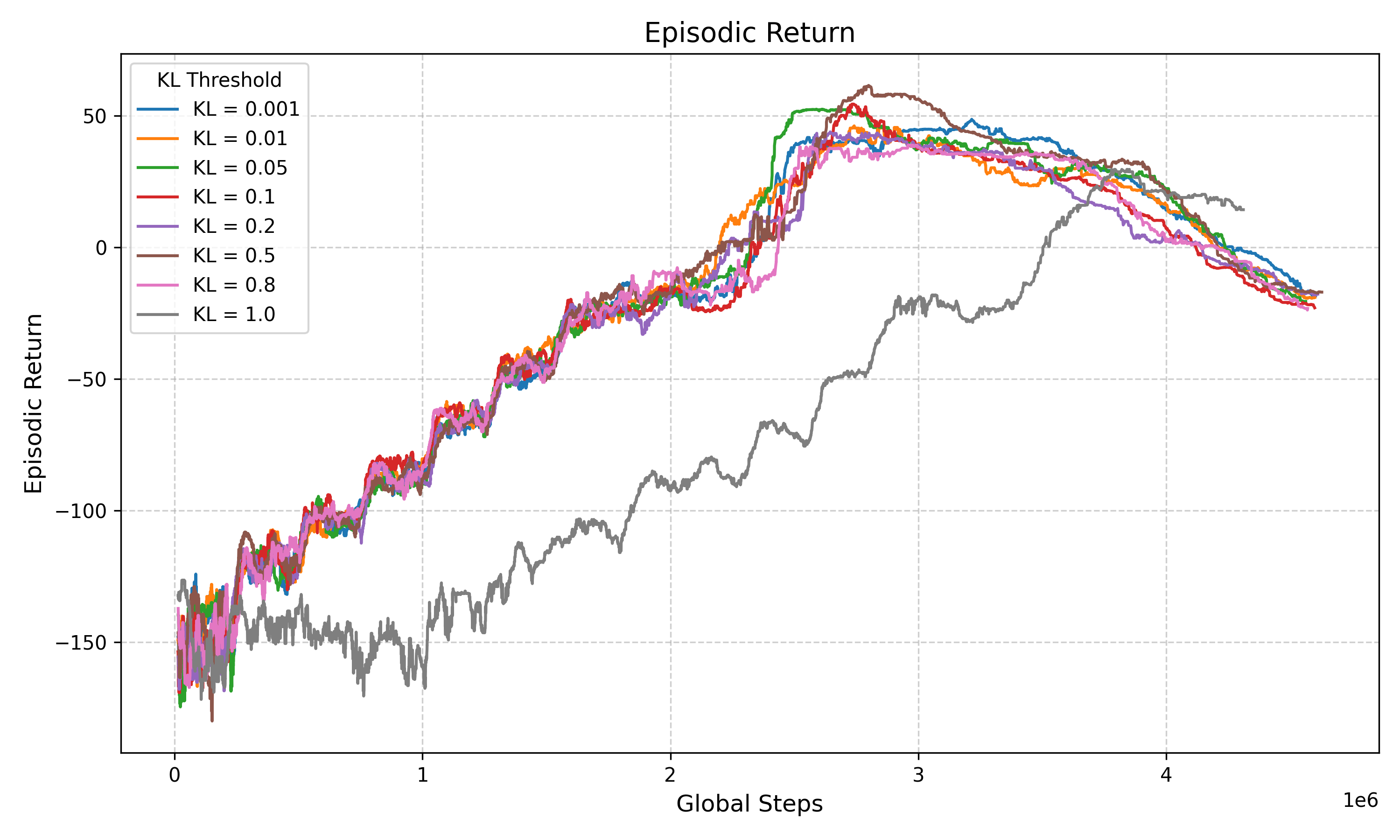}
    \caption{\textbf{Learning curves on LunarLander-v2.} APS with intermediate KL thresholds (e.g., $\delta=0.05$) achieves strong performance with fewer synchronizations compared to lower thresholds.}
    \label{fig:reward-curve}
\end{figure}

\begin{figure}[ht]
    \centering
    \includegraphics[width=0.95\linewidth]{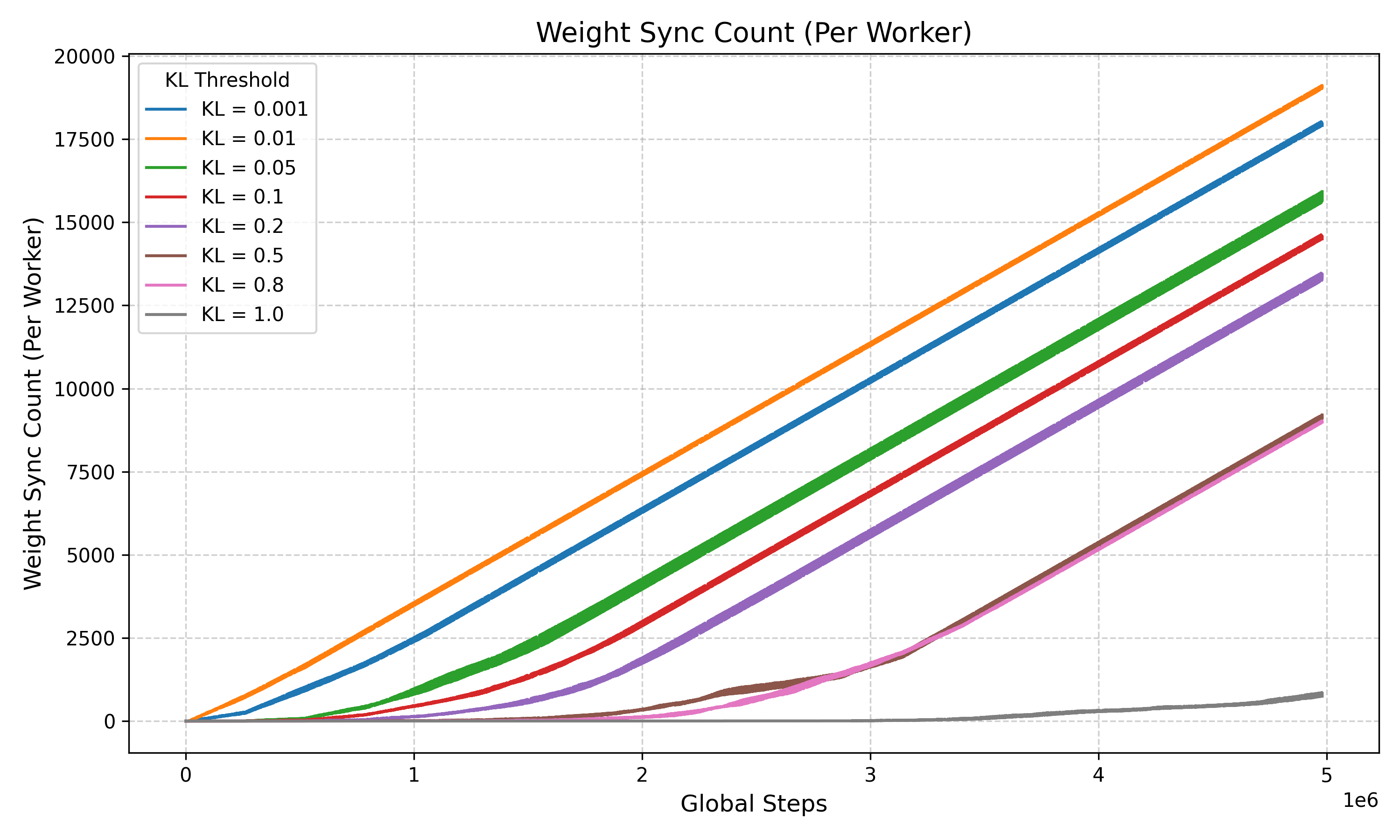}
    \caption{\textbf{Cumulative synchronization count per worker.} Lower KL thresholds result in more frequent weight pulls, while higher thresholds yield computational savings.}
    \label{fig:sync-count}
\end{figure}

\subsection{Discussion}

We find that APS achieves competitive learning across a wide range of KL thresholds, while significantly reducing communication cost. Notably, even with $\delta=0.8$, agents converge effectively, suggesting that occasional stale policies are tolerable in this environment. However, extremely high thresholds (e.g., $\delta=1.0$) may delay convergence due to excessive policy drift.

\section{Future Work}

Future work will extend ClusterEnv along three primary axes: orchestration flexibility, environment richness, and support for asynchronous execution.

First, while ClusterEnv currently targets SLURM-based clusters, adding support for container orchestration platforms such as Kubernetes would enable wider adoption in industry environments. A Kubernetes backend would allow for dynamic scaling, containerized isolation, and compatibility with existing cloud-native workflows.

Second, we aim to extend support to continuous-control and high-dimensional simulation environments, including those with persistent dynamics and non-episodic lifecycles. This requires robust handling of long-lived environment instances and partial resets, which will be incorporated into the ClusterEnv API.

Finally, we plan to support fully asynchronous rollout collection. Although ClusterEnv currently follows a semi-synchronous model, enabling true asynchronous execution across environments would improve throughput under heterogeneous system loads and make better use of compute in large, multi-tenant clusters.

\section{Conclusion}

We presented ClusterEnv, a lightweight, modular interface for distributed reinforcement learning that isolates environment execution from training logic. ClusterEnv enables scalable rollout collection without enforcing a prescriptive learning pipeline, making it compatible with a wide range of algorithms and infrastructures.

Our design is centered on two key contributions: the DETACH architecture, which cleanly separates simulation and learning, and Adaptive Policy Synchronization (APS), a bandwidth-efficient strategy for mitigating policy drift. Together, these components allow DRL practitioners to scale environment interaction across distributed compute with minimal overhead, minimal code changes, and maximal flexibility.

ClusterEnv lowers the barrier to high-throughput distributed RL and provides a clean abstraction that integrates easily into existing research and production workflows. All source code is available at \url{https://github.com/rodlaf/ClusterEnv}.

\bibliographystyle{ieeetr}
\bibliography{ref}

\end{document}